\documentclass[10pt,twocolumn,letterpaper]{article}

\usepackage{iccv} 
\usepackage{times}
\usepackage{epsfig}
\usepackage{graphicx}
\usepackage{amsmath}
\usepackage{amssymb}
\usepackage{color}
\usepackage{algorithmic}
\usepackage{breqn}
\usepackage{booktabs}
\usepackage{fixltx2e} 
\usepackage{multirow}
\usepackage{siunitx}

\usepackage[pagebackref=true,breaklinks=true,letterpaper=true,colorlinks,bookmarks=false]{hyperref}

\iccvfinalcopy 

\def\iccvPaperID{4318} 

\ificcvfinal\fi

\begin{document}

\title{Semi-Supervised Semantic Segmentation with \\ Pixel-Level Contrastive Learning from a Class-wise  Memory Bank}

\author{Inigo Alonso$^1$~~~~~~~~Alberto Sabater$^1$~~~~~~~~David Ferstl$^2$~~~~~~~~Luis Montesano$^{1,}$ ~~~~~~~~Ana C. Murillo$^1$\\
$^1$RoPeRT group, at DIIS - I3A, Universidad de Zaragoza, Spain\\ 
$^2$Magic Leap, Zürich, Switzerland\\ 
$^3$Bitbrain, Zaragoza, Spain\\ 
{\tt\small \{inigo, asabater, montesano, acm\}@unizar.es, dferstl@magicleap.com}

}

\maketitle
\ificcvfinal\thispagestyle{empty}\fi


\newif\ifdraft
\draftfalse


\ificcvfinal
 \newcommand{\INIGO}[1]{}
 \newcommand{\inigo}[1]{}
 \newcommand{\DF}[1]{}
 \newcommand{\df}[1]{}
 \newcommand{\ANA}[1]{}
 \newcommand{\ana}[1]{}
 \newcommand{\LUIS}[1]{}
 \newcommand{\luis}[1]{}
\else
 \newcommand{\INIGO}[1]{{\color{orange}{\bf Inigo: #1}}}
 \newcommand{\inigo}[1]{{\color{orange} #1}}
 \newcommand{\DF}[1]{{\color{cyan}{\bf David: #1}}}
 \newcommand{\df}[1]{{\color{cyan} #1}}
 \newcommand{\ANA}[1]{{\color{purple}{\bf Ana: #1}}}
 \newcommand{\ana}[1]{{\color{purple} #1}}
 \newcommand{\LUIS}[1]{{\color{blue}{\bf Luis: #1}}}
 \newcommand{\luis}[1]{{\color{blue} #1}}
\fi

\newcommand{\comment}[1]{}
\newcommand{\parag}[1]{\paragraph{#1}}
\renewcommand{\floatpagefraction}{.99}

\newcommand{\cF}{\mathcal{F}}

\newcommand{\bA}{\mathbf{A}}
\newcommand{\bC}{\mathbf{C}}
\newcommand{\bI}{\mathbf{I}}
\newcommand{\bH}{\mathbf{H}}
\newcommand{\bR}{\mathbf{R}}
\newcommand{\bM}{\mathbf{M}}
\newcommand{\bO}{\mathbf{O}}

\newcommand{\real}{\mathbb{R}}

\newcommand{\radius}{\mathbf{r}}

\newcommand{\OURS}[0]{\textbf{OURS}}

\newcommand\norm[1]{\left\lVert#1\right\rVert}

\newcommand{\colvecTwo}[2]{\ensuremath{
		\begin{bmatrix}{#1}	\\	{#2}	\end{bmatrix}
}}
\newcommand{\colvec}[3]{\ensuremath{
		\begin{bmatrix}{#1}	\\	{#2}	\\	{#3} \end{bmatrix}
}}
\newcommand{\colvecFour}[4]{\ensuremath{
		\begin{bmatrix}{#1}	\\	{#2}	\\	{#3} \\	{#4}	\end{bmatrix}
}}

\newcommand{\rowvecTwo}[2]{\ensuremath{
		\begin{bmatrix}{#1}	&	{#2}	\end{bmatrix}
}}
\newcommand{\rowvec}[3]{\ensuremath{
		\begin{bmatrix}{#1} &	{#2}	&	{#3} \end{bmatrix}
}}
\newcommand{\rowvecFour}[4]{\ensuremath{
		\begin{bmatrix}{#1}	&	{#2}	&	{#3} &	{#4}	\end{bmatrix}
}}

\newcommand{\fig}[1]{figure~\ref{#1}}
\newcommand{\tab}[1]{table~\ref{#1}}
\newcommand{\Fig}[1]{Figure~\ref{#1}}
\newcommand{\Sec}[1]{Section~\ref{#1}}
\newcommand{\Tab}[1]{Table~\ref{#1}}
\newcommand{\equ}[1]{(Equation~\ref{#1})}

\begin{abstract} 

This work presents a novel approach for semi-supervised semantic segmentation. The key element of this approach is our contrastive learning module that enforces the segmentation network to yield similar pixel-level feature representations for same-class samples across the whole dataset. 
To achieve this, we maintain a memory bank continuously updated with relevant and high-quality feature vectors from labeled data.
In an end-to-end training, the features from both labeled and unlabeled data are optimized to be similar to same-class samples from the memory bank. 
Our approach outperforms the current state-of-the-art for semi-supervised semantic segmentation and semi-supervised domain adaptation on well-known public benchmarks, with larger improvements on the most challenging scenarios, i.e., less available labeled data.  \url{https://github.com/Shathe/SemiSeg-Contrastive}
\end{abstract}

\section{Introduction}

Semantic segmentation consists in assigning a semantic label to each pixel in an image. It is an essential computer vision task for scene understanding that plays a relevant role in many applications such as medical imaging~\cite{ronneberger2015u} or autonomous driving~\cite{badrinarayanan2017segnet}. 
As for many other computer vision tasks, deep convolutional neural networks have shown significant improvements in semantic segmentation~\cite{badrinarayanan2017segnet, jegou2017one, alonso2020MininetV2}. All these examples follow supervised approaches requiring a large set of annotated data to generalize well. However, the availability of labels is a common bottleneck in supervised learning, especially for tasks such as semantic segmentation, which require expensive per-pixel annotations. 

\begin{figure}[!tb]
\centering
\includegraphics[width=1\linewidth]{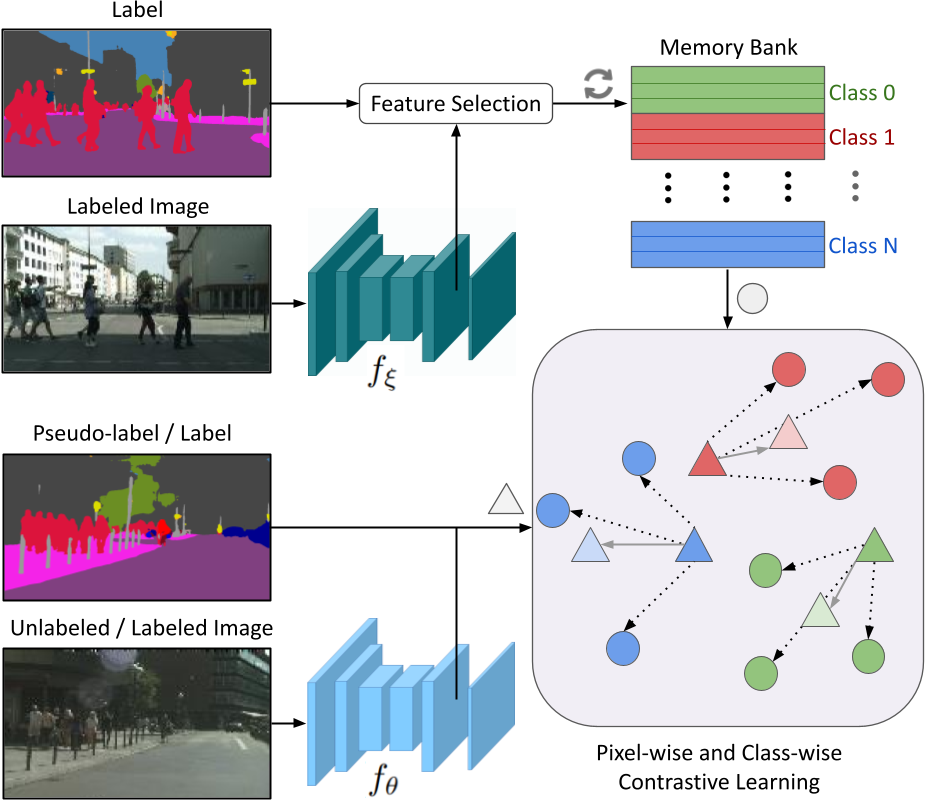}
\caption{
\textbf{Proposed contrastive learning module overview.} At each training iteration, the teacher network $f_{\xi}$ updates the feature memory bank with a subset of selected features from labeled samples.
Then, the student network $f_{\theta}$ extracts features $\triangle$ from both labeled and unlabeled samples, 
which are optimized to be similar to same-class features from the memory bank 
\scriptsize \textbigcircle \small. \vspace{-3mm}
}\label{fig:introduction}
\end{figure}

Semi-supervised learning assumes that only a small subset of the available data is labeled. It tackles this limited labeled data issue by extracting knowledge from unlabeled samples. 
Semi-supervised learning has been applied to a wide range of applications~\cite{van2020survey}, including semantic segmentation~\cite{french2019semi, hung2018adversarial, mittal2019semi}.
Previous semi-supervised segmentation works are mostly based on per-sample entropy minimization~\cite{hung2018adversarial, lee2013pseudo, olsson2021classmix} and per-sample consistency regularization~\cite{french2019semi, tarvainen2017mean, olsson2021classmix}. These segmentation methods do not enforce any type of structure on the learned features to increase inter-class separability across the whole dataset. Our hypothesis is that overcoming this limitation can lead to better feature learning and performance, especially when the amount of available labeled data is low.

This work presents a novel approach for semi-supervised semantic segmentation.
Our approach follows a teacher-student scheme where the main component is a novel representation learning module (\Fig{fig:introduction}). This module is based on positive-only contrastive learning \cite{chen2020exploring, grill2020bootstrap} and enforces the class-separability of pixel-level features across different samples. To achieve this, the teacher network produces feature candidates, only from labeled data, to be stored in a memory bank. Meanwhile, the student network learns to produce similar class-wise features from both labeled and unlabeled data.
The features stored in the memory bank are selected based on their quality and learned relevance for the contrastive optimization. Besides, our module enforces the alignment of unlabeled and labeled data (memory bank) in the feature space, which is another unexploited idea in semi-supervised semantic segmentation. Our main contributions are the following:\vspace{-2mm}
\begin{itemize}
    \item A novel semi-supervised semantic segmentation framework. \vspace{-2mm}
    \item The use of a memory bank for high-quality pixel-level features from labeled data.\vspace{-2mm}
    \item A pixel-level contrastive learning scheme where elements are weighted based on their relevance.\vspace{-2mm}
    \end{itemize}

The effectiveness of our method is demonstrated on well-known  semi-supervised semantic segmentation benchmarks, reaching the state-of-the-art on different set-ups.
Besides, our approach can naturally tackle the semi-supervised domain adaptation task, obtaining state-of-the-art results too. 
In all cases, the improvements upon comparable methods increase with the percentage of unlabeled data.

\section{Related Work}
This section summarizes relevant related work for semi-supervised learning and contrastive learning, with particular emphasis on work related to semantic segmentation.

\subsection{Semi-Supervised Learning}
\paragraph{Pseudo-Labeling}
Pseudo-labeling leverages the idea of creating artificial labels for unlabeled data~\cite{mclachlan1975iterative, scudder1965probability} by keeping the most likely predicted class by an existing model~\cite{lee2013pseudo}.
The use of pseudo-labels is motivated by entropy minimization~\cite{grandvalet2004semi}, encouraging the network to output highly confident probabilities on unlabeled data.
Both pseudo-labeling and direct entropy minimization methods are commonly used in semi-supervised scenarios~\cite{feng2020semiCBC, kalluri2019universal, sohn2020fixmatch, olsson2021classmix} showing great performance.
Our approach makes use of both pseudo-labels and direct entropy minimization.

\paragraph{Consistency Regularization}
Consistency regularization relies on the assumption that the model should be invariant to perturbations, \eg data augmentation, made to the same image. This regularization is commonly applied by using two different methods: distribution alignment~\cite{berthelot2019mixmatch, sajjadi2016regularization, tarvainenweight}, or augmentation anchoring~\cite{sohn2020fixmatch}. 
While distribution alignment enforces the prediction of perturbed and non-perturbed samples to have the same class distribution, augmentation anchoring enforces them to have the same semantic label. 
To produce high-quality non-perturbed class distribution or prediction on unlabeled data, the Mean Teacher method~\cite{tarvainen2017mean}, proposes a teacher-student scheme where the teacher network is an exponential moving average (EMA) of model parameters, producing more robust predictions.


\subsection{Semi-Supervised Semantic Segmentation}
 One common approach for semi-supervised semantic segmentation is to make use of Generative Adversarial Networks (GANs)~\cite{goodfellow2014generative}. Hung \etal~\cite{hung2018adversarial} propose to train the discriminator to distinguish between confidence maps from labeled and unlabeled data predictions. Mittal \etal~\cite{mittal2019semi} make use of a two-branch approach, one branch enforcing low entropy predictions using a GAN approach and another branch for removing false-positive predictions using a Mean Teacher method~\cite{tarvainenweight}.
A similar idea was proposed by Feng \etal~\cite{feng2020semi}, a recent work that introduces Dynamic Mutual Training (DMT). DMT uses two  models and the model's disagreement is used to re-weight the loss. DMT method also followed the multi-stage training protocol from CBC~\cite{feng2020semiCBC}, where pseudo-labels are generated in an offline curriculum fashion.
Other works are based on data augmentation methods for consistency regularization. French \etal~\cite{french2019semi} focus on applying CutOut~\cite{devries2017cutout} and CutMix~\cite{yun2019cutmix}, while Olsson \etal~\cite{olsson2021classmix} propose a data augmentation technique specific for semantic segmentation.


\subsection{Contrastive Learning}
The core idea of contrastive learning~\cite{hadsell2006dimensionality} is to create positive and negative data pairs, to attract the positive and repulse the negative pairs in the feature space. This technique has been used in supervised and self-supervised set-ups.
However, recent self-supervised methods have shown similar accuracy with contrastive learning using positive pairs only by performing similarity maximization with distillation \cite{chen2020exploring, grill2020bootstrap} or redundancy reduction \cite{zbontar2021barlow}.

As for semantic segmentation, these techniques has been mainly used as pre-training~\cite{wang2020dense, xie2020pointcontrast, xie2020propagate}.
Very recently, Wang \etal~\cite{wang2021exploring} have shown improvements in supervised scenarios applying standard contrastive learning in a pixel and region level for same-class supervised samples. Van \etal~\cite{van2021unsupervised} have shown the advantages of contrastive learning in unsupervised set-ups, applying it between features from different saliency masks. Lai \etal \cite{lai2021semi} proposed to use contrastive learning in a self-supervised fashion where positive samples were the same pixel from different views/crops and negative samples were different pixels from a different view, making the model invariant to context information.

In this work, we propose to follow a positive-only contrastive learning based on similarity maximization and distillation \cite{chen2020exploring, grill2020bootstrap}. This way, we boost the performance on semi-supervised semantic segmentation in a simpler and more computationally efficient fashion than standard contrastive learning \cite{wang2021exploring}.
Differently from previous works, our contrastive learning module tackles a semi-supervised scenario aligning class-wise and per-pixel features from both labeled and unlabeled data to features from all over the labeled set that are stored in a memory bank.  
Contrary to previous contrastive learning works \cite{wu2018unsupervised, he2020momentum} that saved image-level features in a memory bank, our memory bank saves  per-pixel features for the different semantic classes.
Besides, as there is not infinite memory for all dataset pixels, we propose to  only save features with the highest quality.  
\vspace{-2mm}


\section{Method}
Semi-supervised semantic segmentation is a per-pixel classification task  where two different sources of data are available: a small set of labeled samples $\mathcal{X}{_l}=\left\{x_{l}, y_{l}\right\}$, where $x_{l}$ are images and $y_{l}$ their corresponding annotations, and a large set of unlabeled samples $\mathcal{X}{_u}=\left\{x_{u}\right\}$.

\begin{figure}[!tb]
\centering
\includegraphics[width=1\linewidth]{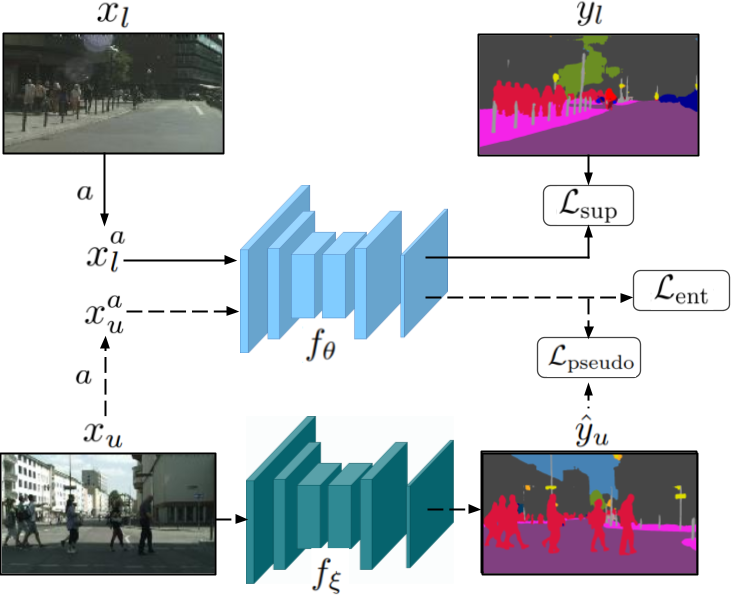}
\caption{
\textbf{Supervised and self-supervised optimization.} The student  $f_{\theta}$ is optimized with $\mathcal{L}_{\mathrm{sup}}$ for labeled data ($x_{l}, y_{l}$). For unlabeled data $x_{u}$, the teacher  $f_{\xi}$ computes the pseudo-labels $\hat{y}_{u}$ that are later used for optimizing $\mathcal{L}_{\mathrm{pseudo}}$ for pairs of augmented samples and pseudo-labels ($x_{u}^{a}$, $\hat{y}_{u}$). Finally, $\mathcal{L}_{\mathrm{ent}}$ is optimized for predictions from $x_{u}^{a}$. } 
\label{fig:self-training}
\vspace{-2mm}
\end{figure}

To tackle this task, we propose to use a teacher-student scheme. The teacher network $f_{\xi}$ creates robust pseudo-labels from unlabeled samples and memory bank entries from labeled samples to teach the student network $f_{\theta}$ to improve its segmentation performance. \vspace{-4mm}

\paragraph{Teacher-student scheme.} The learned weights $\theta$ of the student network $f_{\theta}$ are optimized using the following loss:

\begin{equation}
\label{eq:approach}
\mathcal{L} = \lambda_{sup}\mathcal{L}_{\mathrm{sup}} + \lambda_{pseudo}\mathcal{L}_{\mathrm{pseudo}} + \lambda_{ent}\mathcal{L}_{\mathrm{ent}} + \lambda_{contr}\mathcal{L}_{\mathrm{contr}}.
\end{equation}
$\mathcal{L}_{\mathrm{sup}}$ is a supervised learning loss on labeled samples (\Sec{sec:supervised}). $\mathcal{L}_{\mathrm{pseudo}}$ and $\mathcal{L}_{\mathrm{ent}}$ tackle pseudo-labels (\Sec{sec:pseudolabels}) and entropy minimization (\Sec{sec:entropy}) techniques, respectively, where  pseudo-labels are generated by the teacher network $f_{\xi}$. Finally, $\mathcal{L}_{\mathrm{contr}}$ is our proposed positive-only contrastive learning loss (\Sec{sec:contrastive}).

Weights $\xi$ of the teacher network $f_{\xi}$ are an exponential moving average of weights $\theta$ of the student network $f_{\theta}$ with a decay rate $\tau \in[0,1]$. The teacher model provides more accurate and robust predictions \cite{tarvainen2017mean}.
Thus, at every training step, the teacher network $f_{\xi}$ is not optimized by a gradient descent but updated as follows:\vspace{-1mm}
\begin{equation}
\label{eq:ema}
 \xi = \tau \xi+(1-\tau) \theta. \vspace{-2mm}
\end{equation}

\begin{figure*}[!tb]
\centering
\includegraphics[width=1\linewidth]{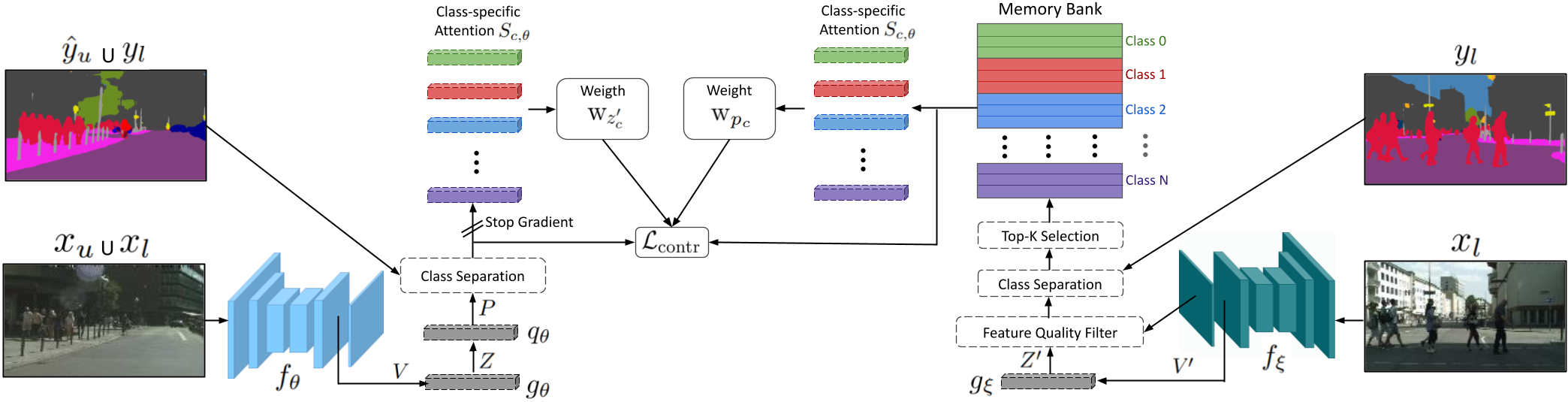}
\caption{
\textbf{Contrastive learning optimization.} At every iteration, features are extracted by $f_{\xi}$ from labeled samples (see right part). These features are projected, filtered by their quality, and then, ranked to finally only store the highest-quality features into the memory bank.
Concurrently, feature vectors from input samples extracted by $f_{\theta}$ are fed to the projection and prediction heads (see left part). Then, feature vectors are passed to a self-attention module in a class-wise fashion, getting a per-sample weight. Finally, input feature vectors are enforced to be similar to same-class features from the memory bank.}
\label{fig:contrastive}
\end{figure*}

\subsection{Supervised Segmentation: $\mathcal{L}_{\mathrm{sup}}$}
\label{sec:supervised}
Our supervised semantic segmentation optimization, applied to the labeled data $\mathcal{X}{_l}$, follows the standard optimization with the weighted cross-entropy loss. Let $\mathcal{H}$ be the weighted cross-entropy loss between two lists of $N$ per-pixel class probability distributions $y_1, y_2$:
\begin{equation}
\label{eq:crossentropy}
\mathcal{H}(y_1, y_2) = - \frac{1}{N}\sum_{n=1}^N\sum_{c=1}^{C} y_2^{(n,c)} \log (y_1^{(n,c)}) \alpha^{c} \beta^{n},
\end{equation}

\noindent where $C$ is the number of classes to classify, $N$ is the number of elements, \ie pixels in $y_1$, $\alpha^{c}$ is a per-class weight, and, $\beta^{n}$ is a per-pixel weight. Specific values of $\alpha^{c}$ and $\beta^{n}$ are detailed in \Sec{sec:implementation}. The supervised loss (see top part of \Fig{fig:self-training}) is defined as follows: 
\begin{equation}
\label{eq:supervised}
\mathcal{L}_{\mathrm{sup}}= \mathcal{H}\left(f_{\theta}\left(x_{l}^{a}\right), y_{l}\right), 
\end{equation}

\noindent where $x_{l}^{a}$ is a weak augmentation of $x_{l}$ (see \Sec{sec:implementation} for augmentation details). 

\subsection{Learning from Pseudo-labels: $\mathcal{L}_{\mathrm{pseudo}}$}
\label{sec:pseudolabels}

The key to the success of semi-supervised learning is to learn from unlabeled data. One idea our approach exploits is to learn from pseudo-labels.
In our case, pseudo-labels are generated by the teacher network $f_{\xi}$ (see \Fig{fig:self-training}). 
For every unlabeled sample $x_{u}$, the pseudo-labels $\hat{y}_{u}$ are computed following this equation:\vspace{-1mm}
\begin{equation}
\label{eq:psuedolabels}
\hat{y}_{u} = \arg\max f_{\xi}\left(x_{u}\right),
\end{equation} 
\noindent where $f_{\xi}$ predicts a class probability distribution. 
Note that pseudo-labels are computed at each training iteration.

Consistency regularization is introduced with augmentation anchoring, \ie computing different data augmentation for each sample $x_{u}$ on the same batch, helping the model to converge to a better solution~\cite{sohn2020fixmatch}. 
The pseudo-labels loss for unlabeled data $\mathcal{X}{_u}$ is calculated by the cross-entropy:\vspace{-2mm}
\begin{equation}
\label{eq:self-supervised}
\mathcal{L}_{\mathrm{pseudo}}= \frac{1}{A}\sum_{a=1}^{A} \mathcal{H}\left(f_{\theta}\left(x_{u}^{a}\right), \hat{y}_{u}\right),
\end{equation}
where $x_{u}^{a}$ is a strong augmentation of $x_{u}$ and $A$ is the number of augmentations we apply to sample $x_{u}$ (see \Sec{sec:implementation} for augmentation details). 

\subsection{Direct Entropy Minimization: $\mathcal{L}_{\mathrm{ent}}$}
\label{sec:entropy}
Direct entropy minimization is applied on the class distributions predicted by the student network from unlabeled samples $x_{u}$ as a regularization loss: \vspace{-2mm}
\begin{equation}
\label{eq:entropy}
\mathcal{L}_{ent} = - \frac{1}{A}\frac{1}{N}\sum_{a=1}^{A}\sum_{n=1}^N\sum_{c=1}^{C} f_{\theta}\left(x_{u}^{a,n,c} \right)\log f_{\theta}\left(x_{u}^{a, n,c}\right), 
\end{equation}
where $C$ is the number of classes to classify, $N$ is the number of pixels and $A$ is the number of augmentations.

\subsection{Contrastive Learning: $\mathcal{L}_{\mathrm{contr}}$}
\label{sec:contrastive}

\Fig{fig:contrastive} illustrates our proposed contrastive optimization inspired by positive-only contrastive learning works based on similarity maximization and distillation \cite{chen2020exploring, grill2020bootstrap}. In our approach, a memory bank is filled with high-quality feature vectors from the teacher $f_{\xi}$ (right part of \Fig{fig:contrastive}). Concurrently, the student $f_{\theta}$ extracts feature vectors from either $\mathcal{X}{_l}$ or $\mathcal{X}{_u}$. In a per-class fashion, every feature is passed through a simple self-attention module that serves as per-feature weighting in the contrastive loss. Finally, the loss enforces the weighted feature vectors from the student to be similar to feature vectors from the memory bank. 
As the memory bank contains high-quality features from all labeled samples, the contrastive loss helps to create a better class separation in the feature space across the whole dataset as well as aligning the unlabeled data distribution with the labeled data distribution.  \vspace{-2mm}

\paragraph{Optimization.}
Let $f_{\theta^-}$ be the student network without the classification layer and $\left\{{x, y}\right\}$ a training sample either from $\left\{\mathcal{X}{_l},\mathrm{Y}{_l}\right\}$ or $\{\mathcal{X}{_u}, \mathrm{\hat{Y}}{_u}\}$. The first step is to extract all feature vectors: $V = f_{\theta^-}(x)$. The feature vectors $V$ are then fed to a projection head, $Z=g_{\theta}(V)$, and a prediction head, $P=q_{\theta}(Z)$, following~\cite{chen2020exploring, grill2020bootstrap}, where $g_{\theta}$ and $q_{\theta}$ are two different Multi-Layer Perceptrons (MLPs).
Next, $P$ is grouped by the different semantic classes in $y$. 

Let $P_c=\left\{p_{c}\right\}$ be the set of prediction vectors from $P$ of a  class $c$. Let $Z_c'=\left\{z_{c}'\right\}$ be the set of projection vectors of class $c$ obtained by the teacher, $Z' = g_{\xi}(f_{\xi^-}(x))$ from the labeled examples stored in the memory bank.

Next, we learn which feature vectors ($p_{c}$ and $z_{c}'$) are beneficial for the contrastive task, by assigning per-feature learned weights \equ{eq:re-weighting} that will serve as a weighting factor \equ{eq:weighted-distance} for the contrastive loss function \equ{eq:contrastive-loss}.
These per-feature weights are computed using class-specific attention modules $S_{c, \theta}$ (see \Sec{sec:implementation} for further details) that generate a single value ($\mathrm{w} \in [0, 1]$) for every $z_{c}'$ and $p_{c}$ feature. 
Following~\cite{sun2020learning} we L1 normalize these weights to prevent converging to the trivial all-zeros solution. For the prediction vectors $P_c$ case, the weights $\mathrm{w}{_{p_c}}$ are then computed as follows: \vspace{-1mm}
\begin{equation}
\label{eq:re-weighting}
\mathrm{w}{_{p_c}} = \frac{N_{P_c}}{\sum_{\mathrm{p_i} \in P_c} S_{c, \theta}(p_{i})} S_{c, \theta}(p_{c}),
\end{equation}
where $N_{P_c}$ is the number of elements in $P_c$. Equation~\ref{eq:re-weighting} is used to compute $\mathrm{w}{_{z_c'}}$ too, changing $Z_c'$ and $z_c'$ for $P_c$ and $p_c'$. 

The contrastive loss enforces prediction vectors $p_c$ to be similar to projection vectors $z_c'$ as in \cite{chen2020exploring, grill2020bootstrap} (in our case, projection vectors are in the memory bank). For that, we use the cosine similarity as the similarity measure ${C}$: \vspace{-2mm}
\begin{equation}
\label{eq:distance}
\mathcal{C}(p_c, z_c') = \frac{\left\langle p_c, z_c'\right\rangle}{\left\|p_c\right\|_{2} \cdot\left\|z_c'\right\|_{2}}, 
\end{equation}
where, the weighted distance between predictions and memory bank entry is computed by: \vspace{-1mm}
\begin{equation}
\label{eq:weighted-distance}
\mathcal{{D}}(p_c, z_c') = \mathrm{w}{_{p_c}} \mathrm{w}{_{z_c'}} (1 - \mathcal{C}(p_c, z_c')), 
\end{equation}
and, our contrastive loss is computed as follows: \vspace{-2mm}
\begin{equation}
\label{eq:contrastive-loss}
\mathcal{L}_{contr} = \frac{1}{C} \frac{1}{N_{p_c}} \frac{1}{N_{z_c'}} \sum_{c=1}^{C} \sum_{{p_c} \in P_c}\sum_{{z_c'} \in Z_c'} \mathcal{D}(p_c, z_c').
\end{equation}

\paragraph{Memory Bank.}
The memory bank is the data structure that maintains the target feature vectors $z_c'$, $\psi$ for each class $c$, used in the contrastive loss. As there is not infinite space for saving all pixels of the labeled data, we propose to store only a subset of the feature vectors from labeled data with the highest quality.
As shown in \Fig{fig:contrastive}, the memory bank is updated on every training iteration with a subset of $z_c' \in Z'$ generated by the teacher. To select what subset of $Z'$ is included in the memory bank, we first perform a Feature Quality Filter (FQF), where we only keep features that lead to an accurate prediction when the classification layer is applied, $y=\arg\max f_{\xi}(x_l)$, having confidence higher than a threshold, $f_{\xi}(x_l) > \phi$. 
The remaining $Z'$ are grouped by classes $Z_c'$. Finally, instead of picking randomly a subset of every $Z_c'$ to update the memory bank, we make use of the class-specific attention modules $S_{c, \xi}$. We get ranking scores $R_c = S_{c, \xi}(Z_c')$ to sort $Z_c'$ and we update the memory bank only with the top-K highest-scoring vectors. The memory bank is a First In First Out (FIFO) queue per class for computation and time efficiency. This way it maintains recent high-quality feature vectors in a very efficient fashion computation-wise and time-wise. Detailed information about the hyper-parameters is included in \Sec{sec:implementation}.

\vspace{-1.5mm}

\section{Experiments}
\label{sec:experiments}
This section describes evaluation set-up and the evaluation of our method on different benchmarks for semi-supervised semantic segmentation, including a semi-supervised domain adaptation, and a detailed ablation study. 
\vspace{-3.5mm}
\subsection{Datasets}
\begin{itemize}
    \item Cityscapes~\cite{cordts2016cityscapes}. It is a real urban scene dataset composed of $2975$ training and $500$ validation samples, with $19$ semantic classes. \vspace{-1.5mm}
    \item PASCAL VOC 2012~\cite{everingham2010pascal}. It is a natural scenes dataset with $21$ semantic classes. The dataset has $10582$ and $1449$ images for training and validation respectively.\vspace{-1.5mm}
    
    \item GTA5~\cite{richter2016playing}. It is a synthetic dataset captured from a video game with realistic urban-like scenarios with $24966$ images in total. The original dataset provides $33$ different categories but, following~\cite{wang2020alleviating}, we only use the $19$ classes that are shared with Cityscapes.\vspace{-1mm}
\end{itemize}

\subsection{Implementation details}
\label{sec:implementation}
\paragraph{Architecture.} We use DeepLab networks~\cite{chen2017deeplab} in our experiments. For the ablation study and most benchmarking experiments, DeepLabv2 with a ResNet101 backbone is used to have similar settings to previous works~\cite{olsson2021classmix, feng2020semiCBC, hung2018adversarial, mittal2019semi}. DeepLabv3+ with Resnet50 backbone is also used to equal comparison with~\cite{mendel2020semi,lai2021semi}. $\tau$ is set from 0.995 to 1 during training in \equ{eq:ema}. 

The prediction and projection heads follow~\cite{grill2020bootstrap}: Linear (256) $\rightarrow$ BatchNorm~\cite{ioffe2015batch} $\rightarrow$ Relu~\cite{nair2010rectified} $\rightarrow$ Linear (256).
The proposed class-specific attention modules follow a similar architecture: Linear (256 )$\rightarrow$ BatchNorm $\rightarrow$ LeakyRelu~\cite{maas2013rectifier} $\rightarrow$ Linear (1) $\rightarrow$ Sigmoid. We use $2\times N_{classes}$ attention modules since they are used in a class-wise fashion. In particular, two modules per class are used because we have different modules for projection or prediction feature vectors.

\vspace{-3mm}
\paragraph{Optimization.}
For all experiments, we train for 150K iterations using the SGD optimizer with a momentum of $0.9$. The learning rate is set to $2\times10^{-4}$ for DeepLabv2 and $4\times10^{-4}$ for DeepLabv3+ with a poly learning rate schedule. For the Cityscapes and GTA5 datasets, we use a crop size of $512\times 512$ and batch sizes of 5 and 7 for Deeplabv2 and Deeplabv3+, respectively. For Pascal VOC, we use a crop size of $321\times 321$ and batch sizes of 14 and 20 for Deeplabv2 and Deeplabv3+, respectively.
Cityscapes images are resized to $512\times 1024$ before cropping when Deeplabv2 is used for a fair comparison with~\cite{olsson2021classmix, feng2020semiCBC, hung2018adversarial, mittal2019semi}.
The different loss weights in \equ{eq:approach} are set as follows for all experiments: $\lambda_{sup} = 1$, $\lambda_{pseudo} = 1$, $\lambda_{ent} = 0.01$, $\lambda_{contr} = 0.1$.
An exception is made for the first 2K training iterations where $\lambda_{contr}=0$ and $\lambda_{pseudo} = 0$ to make sure predictions have some quality before being taken into account.
Regarding the per-pixel weights ($\beta^{n}$) from $\mathcal{H}$ in \equ{eq:crossentropy}, we set it to 1 for $\mathcal{L}_{\mathrm{sup}}$. For $\mathcal{L}_{\mathrm{pseudo}}$, we follow~\cite{feng2020semiCBC} weighting each pixel with its corresponding pseudo-label confidence with a sharpening operation, $f_{\xi}\left(x_{u}\right)^s$, where we set $s=6$.
As for the per-class weights $\alpha^{c}$ in \equ{eq:crossentropy}, we perform a class balancing for the Cityscapes and GTA5 datasets by setting $\alpha^{c}= \sqrt{\frac{f_{m}}{f_c}} $ with $f_{c}$ being the frequency of class $c$ and $f_{m}$ the median of all class frequencies. 
In semi-supervised settings the amount of labels, $\mathrm{Y}_{l}$, is usually small. 
For a more meaningful estimation, we compute these frequencies not only from $\mathrm{{Y}}{_l}$ but also from $\mathrm{\hat{Y}}{_u}$. 
For the Pascal VOC we set $\alpha^{c}= 1$ as the class balancing does not have a beneficial effect.

\vspace{-3mm}
\paragraph{Other details.} 
DeepLab's output resolution is $\times8$ lower than the input resolution. For feature comparison during training, we keep the output resolution and downsample the labels reducing memory requirements and computation.

The memory bank size is fixed to $\psi = 256$ vectors per class (see \Sec{sec:ablation} for more details). The confidence threshold $\phi$ for accepting features is set to $0.95$. The number of vectors added to the memory bank at each iteration, for each image, and for each class is set as $\max(1, \frac{\psi}{|\mathcal{X}{_l}|})$, where $|\mathcal{X}{_l}|$ is the number of labeled samples.
 
A single NVIDIA Tesla V100 GPU is used for all experiments. All our reported results are the mean of three different runs with different labeled/unlabeled data splits. 

Following \cite{olsson2021classmix, tarvainen2017mean}, the segmentation is performed with the student $f_{\theta}$ in the experimental validation, although the teacher would lead to a slightly better performance \cite{sohn2020fixmatch}.
\vspace{-3mm}
\paragraph{Data augmentation.}

We use two different augmentation set-ups, a weak one for labeled samples and a strong set-up for unlabeled samples, following \cite{olsson2021classmix}  with minor modifications (Table \ref{tab:transformation_distributions} describes the followed data augmentation scheme in our method). Besides, we set $A=2$ \equ{eq:self-supervised} as the number of augmentations for each sample. \vspace{-1.5mm}

\begin{table}[ht]
    \caption{Strong and weak data augmentation set-ups}
    \footnotesize
    \centering
    \begin{tabular}{l c l} \toprule
        Parameter & Weak & Strong \\ \midrule
        Flip probability & $0.50$ & $0.50$ \\
        Resize $\times[0.75, 1.75$] probability  & $0.50$ & $0.80$\\
        Color jittering probability & $0.20$ & $0.80$ \\
        Brightness adjustment max intensity  & $0.15$ & $0.30$ \\
        Contrast adjustment max intensity & $0.15$ &$0.30$ \\
        Saturation adjustment max intensity  & $0.075$  & $0.15$ \\
        Hue adjustment max intensity  & $0.05$ & $0.10$\\
        Gaussian blurring probability  & $0$ & $0.20$ \\
        ClassMix probability  & $0.20$ & $0.80$ \\ 
\bottomrule
    \end{tabular} 
    \label{tab:transformation_distributions}
\end{table}

\vspace{-4mm}
\subsection{Benchmark Experiments}

Following experiments compare our method with state-of-the-art methods in different semi-supervised scenarios. \vspace{-3mm}

\vspace{-5mm}
\subsubsection{Semi-supervised Semantic Segmentation}

\begin{figure*}[!tb]
\centering
\includegraphics[width=1\linewidth]{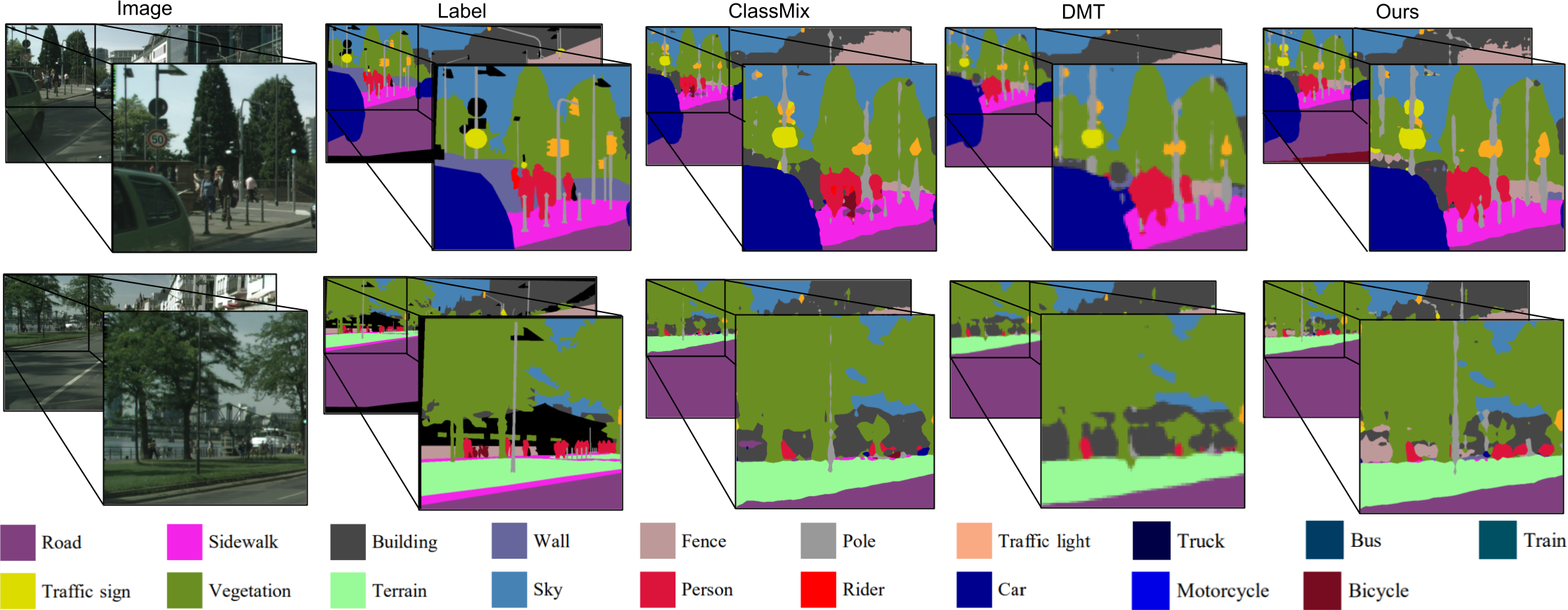}
\caption{Qualitative results on Cityscapes. Models are trained with $\frac{1}{8}$ of the labeled data using Deeplabv2 with ResNet-101. From left to right: Image, manual annotations, ClassMix~\cite{olsson2021classmix}, DMT~\cite{feng2020semi}, our approach.
}
\label{fig:cityscapes-results}
\end{figure*}

\begin{table}[!tb]
\label{table:city}
\centering 
\resizebox{0.49\textwidth}{!}{

\begin{tabular}{l|ccc|c} \toprule
method & 1/30 & 1/8 & 1/4 & FS  \\

\midrule
\multicolumn{5}{@{}p{6.8cm}}{\small{\textit{Architecture}: Deeplabv2 with ResNet-101 backbone}
}\\

\midrule

Adversarial~\cite{hung2018adversarial}+ &---  & $58.8$ (-$7.6$) & $62.3$ (-$4.1$)& $66.4$ \\ 
s4GAN~\cite{mittal2019semi}* & ---  & $59.3$ (-$6.7$)& $61.9$ -($4.9$)& $66.0$\\ 
French \etal~\cite{french2019semi}* & $51.2$  (-$16.3$)   &  $60.3$  (-$7.2$)   & $63.9$  (-$3.6$) & $67.5$\\
CBC~\cite{feng2020semiCBC}+ & $48.7$  (-$18.2$) & $60.5$ (-$6.4$)  & $64.4$  (-$2.5$)& $66.9$  \\
ClassMix~\cite{olsson2021classmix}+ & $54.1$ (-$12.1$)  &  $61.4$  (-$4.8$)& $63.6$ (-$2.6$) &  $66.2$ \\
DMT~\cite{feng2020semi}*+ &  $54.8$ (-$13.4$)  &   $63.0$ (-$5.2$)  &  ---&  $68.2$   \\
\hline
\textbf{Ours}* & $58.0$ (-$8.4$) & $63.0$  (-$3.4$) &$64.8$ (-$1.6$)& $66.4$    \\
\textbf{Ours}+ & $\textbf{59.4 (-7.9)}$  & $\textbf{64.4 (-2.9)}$& $\textbf{65.9 (-1.4)}$  &  $67.3$  \\
  \midrule
\multicolumn{5}{@{}p{7cm}}{\small{\textit{Architecture}: Deeplabv3+ with ResNet-50 backbone
}}\\
 \midrule
 Error-corr~\cite{mendel2020semi}* & --- & $67.4$  (-$7.4$)  & $70.7$ (-$4.1$) &$74.8$ \\
   Lai \etal~\cite{lai2021semi}* & --- &    $69.7$ (-$7.8$)  & $\textbf{72.7}$ (-$4.8$) & $77.5$ \\

\textbf{Ours}* &  $64.9$  (-$9.3$)  & $\textbf{70.1 (-4.1)}$& $71.7 \textbf{ (-2.5)}$  & $74.2$  \\
\bottomrule
\end{tabular}

}
\small{\textit{* ImageNet pre-training, + COCO pre-training} }

\caption{Performance (Mean IoU) for the Cityscapes \textit{val} set for different labeled-unlabeled ratios and, in parentheses, the difference w.r.t. the corresponding fully supervised (FS) result. }
 
\end{table}

\begin{table}[!tb]

\label{table:pascal}
\centering 
\resizebox{0.49\textwidth}{!}{
\begin{tabular}{l|ccc|c} \toprule
method  & 1/50 & 1/20 & 1/8 & FS \\

\midrule
\multicolumn{5}{@{}p{7cm}}{\small{\textit{Architecture}: Deeplabv2 with ResNet-101 backbone}}\\
 \midrule

Adversarial~\cite{hung2018adversarial}+  & $57.2$ (-$17.7$) & $64.7$ (-$10.2$)  & $69.5$  (-$5.4$) &$74.9$  \\
s4GAN~\cite{mittal2019semi}+  & $63.3$ (-$10.3$)  & $67.2$ (-$6.4$)  & $71.4$ (-$2.2$)  &    $73.6$ \\
French \etal~\cite{french2019semi}*  & $64.8$ (-$7.7$)  & $66.5$  (-$6.0$)  & $67.6$ (-$4.9$)  & $72.5$  \\
CBC~\cite{feng2020semiCBC}+  & $65.5$ (-$8.1$)  & $69.3$ (-$4.3$)  & $70.7$ (-$2.9$)  & $73.6$ \\
ClassMix~\cite{olsson2021classmix}+ & $66.2$ (-$7.9$) & $67.8$ (-$6.3$)  &$71.0$ (-$3.1$)   & $74.1$ \\
DMT~\cite{feng2020semi}*+ & $67.2$ (-$7.6$)  & $69.9$ (-$4.9$)  & $\textbf{72.7  (-2.1)}$  & $74.8$  \\  
\hline
\textbf{Ours}* & $65.6$ (-$7.0$)  &$67.8$ (-$4.8$) &  $69.9$ (-$2.7$)  &  $72.6$\\

\textbf{Ours}+ & $\textbf{68.2 (-5.9)}$    &   $\textbf{70.1   (-4.0)}$ & $71.8$ (-$2.3$)  &   $74.1$     \\ 

  \midrule
\multicolumn{5}{@{}p{7cm}}{\small{\textit{Architecture}: Deeplabv3+ with ResNet-50 backbone}}\\
 \midrule  
 Error-corr~\cite{mendel2020semi}* & --- &  --- &  $70.2$ (-$6.1$)  & $76.3$ \\
   Lai \etal~\cite{lai2021semi}* & --- &  --- &  $\textbf{72.4}$ (-$4.1$)  & $76.5$ \\

\textbf{Ours}* & $63.4$ (-$12.5$)   & $69.1$ (-$6.8$) &  $72.0  \textbf{ (-3.9)}$ & $75.9$ \\  
\bottomrule
\end{tabular}
}
\small{\textit{* ImageNet pre-training, + COCO pre-training} }

\caption{Performance (Mean IoU) for the Pascal VOC \textit{val} set for different labeled-unlabeled ratios and, in parentheses, the difference w.r.t. the corresponding fully supervised (FS) result.  
} \vspace{-3mm}
\end{table}

\paragraph{Cityscapes.}
\Tab{table:city} compares different methods on the Cityscapes benchmark for different labeled-unlabeled rates: $\frac{1}{30}$, $\frac{1}{8}$ and, $\frac{1}{4}$. Fully Supervised (FS) scenario, where all images are labeled, is also shown as a reference.
As shown in the table, our approach outperforms the current state-of-the-art by a significant margin. The performance difference is increasing as less labeled data is available, demonstrating the effectiveness of our approach.  
This is particularly important since the goal of semi-supervised learning is to learn with as little supervision as possible. Note that the upper bound for each method is shown in the fully supervised setting (FS). \Fig{fig:cityscapes-results} shows a visual comparison of the top-performing methods on different relevant samples from Cityscapes. 
Note that  Lai \etal~\cite{lai2021semi} have a higher \textit{FS} baseline since they use a higher batch size and crop size among other differences in the set-up.

\vspace{-3mm}
\paragraph{Pascal VOC.}
\Tab{table:pascal} shows the comparison of different methods on the Pascal VOC benchmark, using different labeled-unlabeled rates: $\frac{1}{50}$, $\frac{1}{20}$ and, $\frac{1}{8}$. Our proposed method outperforms previous methods for most of the configurations. Like in the previous benchmark, our method presents larger benefits for the more challenging cases, \ie only a small fraction of data is labeled ($\frac{1}{50}$). This demonstrates that the proposed approach is especially effective to learn from unlabeled data.

\begin{table}[!b]

\label{table:domainadaptation}
\centering 
\small
\begin{tabular}{c|ccc|c} \toprule
 City & ASS~\cite{wang2020alleviating}& Liu \etal~\cite{liu2021domain} & \textbf{Ours}   & \textbf{Ours} \\
 Labels & \multicolumn{3}{c|}{with domain adaptation} &  no adaptation \\

\midrule
$1/30$ & $54.2$  &  $55.2$      &   $\textbf{59.9}$  &  $58.0$        \\ 
$1/15$&  $56.0$   &  $57.0$  &  $\textbf{62.0}$ & $59.9$         \\ 
$1/6$&  $60.2$  &   $60.4$    &  $\textbf{64.2}$  &  $63.7$     \\ 
$1/3$ &  $64.5$   &  $64.6$  &   $\textbf{65.6}$   &  $65.1$    \\ 
\bottomrule

\end{tabular}

\caption{Mean IoU in Cityscapes \textit{val} set. Central columns evaluate the semi-supervised domain adaptation task (GTA5 $\rightarrow$ Cityscapes). The last column evaluates a semi-supervised setting in Cityscapes (no adaptation). 
Different labeled-unlabeled ratios for Cityscapes are compared.
All methods use ImageNet pre-trained Deeplabv2 with ResNet-101 backbone. 
}
\end{table}
 \vspace{-3mm}
\subsubsection{Semi-supervised Domain Adaptation}
Semi-supervised domain adaptation for semantic segmentation differs from the semi-supervised set-up in the availability of labeled data from another domain. That is, apart from having $\mathcal{X}{_l}=\left\{x_{l}, y_{l}\right\}$ and $\mathcal{X}{_u}=\left\{x_{u}\right\}$ from the target domain, a large set of labeled data from another domain is also available: $\mathcal{X}{_d}=\left\{x_{d}, y_{d}\right\}$.
 
Our method can naturally tackle this task by evenly sampling from both $\mathcal{X}{_l}$ and $\mathcal{X}{_d}$ as our labeled data when optimizing $\mathcal{L}_{\mathrm{sup}}$ and $\mathcal{L}_{\mathrm{contr}}$. However, the memory bank only stores features from the target domain $\mathcal{X}{_l}$. 
In this way, 
both the features from unlabeled data $\mathcal{X}{_u}$,
and the features from the other domain $\mathcal{X}{_d}$ are aligned with those from $\mathcal{X}{_l}$.

Following ASS~\cite{wang2020alleviating, liu2021domain}, we take the GTA5 dataset as $\mathcal{X}{_d}$, where all elements are labeled, and the Cityscapes is the target domain consisting of a small set of labeled data $\mathcal{X}{_l}$ and a large set of unlabeled samples $\mathcal{X}{_u}$. \Tab{table:domainadaptation} compares the results of our method with previous  methods \cite{wang2020alleviating, liu2021domain} where all methods use ImageNet pre-training.
For reference, we also show the results of our approach with no adaptation, \ie only training on the target domain Cityscapes, as we do for the semi-supervised set up from the previous
experiment (\Tab{table:city}). We can see that our approach benefits from the use of the other domain data (GTA5), especially where there is little labeled data available ($\frac{1}{30}$), as it could be expected. 
Our method outperforms ASS by a large margin in all the different set-ups. As in previous experiments, our improvement is more significant when the amount of available labeled data is smaller.

\subsection{Ablation Experiments}
 \label{sec:ablation}
The following experiments study the impact of the different components of the proposed approach. The evaluation is done on the Cityscapes data, since it provides more complex scenes compared to Pascal VOC. We select the challenging labeled data ratio of $\frac{1}{30}$. 
 
\begin{table}[!tb]
\label{tab:ablation}
\begin{center}
\setlength{\tabcolsep}{5pt}
\begin{tabular}{cccc|c}
\hline 
$\mathcal{L}_{\mathrm{sup}}$ & $\mathcal{L}_{\mathrm{pseudo}}$ & $\mathcal{L}_{\mathrm{ent}}$ &  $\mathcal{L}_{\mathrm{contr}}$&   mIoU  \\ \hline
\checkmark &  &  &  &     $49.5$   \\
\checkmark & \checkmark &  &  &     $56.7$   \\
\checkmark &  &  \checkmark &  &     $52.2$   \\
\checkmark &  &  &  \checkmark &     $54.4$   \\

\checkmark & \checkmark & \checkmark  &  &    $57.4$  \\
\checkmark &  \checkmark &  &   \checkmark &     $59.0$   \\
\checkmark &  &  \checkmark  &  \checkmark &   $57.3$  \\

\hline
\checkmark & \checkmark & \checkmark  & \checkmark  & $59.4$ \\
\hline 

\end{tabular}

\end{center}

\caption{Ablation study on the different losses included \equ{eq:approach}. Mean IoU obtained on Cityscapes benchmark ($\frac{1}{30}$ available labels, Deeplabv2-ResNet101 COCO pre-trained). \vspace{-3mm}
}
\end{table}

\vspace{-3mm}

\paragraph{Losses impact.} \Tab{tab:ablation} shows the impact of every loss used by the proposed method.
We can observe that the four losses are complementary, getting a $10$ mIoU increase over our baseline model, using only the supervised training when $\frac{1}{30}$ of the Cityscapes labeled data is available. Note that our proposed contrastive learning module $\mathcal{L}_{\mathrm{contr}}$ is able to get $54.32$ mIoU even without any other complementary loss, which is the previous state-of-the-art for this set-up (see \Tab{table:city}). Adding the $\mathcal{L}_{\mathrm{pseudo}}$ significantly improves the performance and then, adding $\mathcal{L}_{\mathrm{ent}}$ regularization loss gives a little extra performance gain. 

Note that at testing time, our approach only uses the student network $f_{\theta}$, adding zero additional computational cost. At training time, for the experiment of \Tab{tab:ablation} having an input resolution of $512\times512$ with a forward pass cost of 372.04 GFLOPs, our method performs $1151.19$ GFLOPs for one training step using one labeled image and one unlabeled image,  compared to the $1488.16$ GFLOPs from  \cite{feng2020semi} or $1116.12$ GFLOPs from \cite{olsson2021classmix}.
The total number of GFLOPs come from $372.04$ for computing labeled image predictions, $372.04$ for the unlabeled image predictions, $372.04$  for computing the pseudo-labels and, $35.07$  for our contrastive module, which mainly include the computation of the prediction and projection heads ($8.59$), the class-specific attention modules  ($15.96$) and, the distance between the input features and memory bank features ($10.52$).

\vspace{-2mm} 
\paragraph{Contrastive learning module.} 
\Tab{tab:lr} shows an ablation on the influence of the contrastive learning module for different values of $\lambda_{contr}$ \equ{eq:approach}.
As expected, if this value is too low, the effect gets diluted, with similar performance as if the proposed loss is not used at all (see \Tab{tab:ablation}). High values are also detrimental, probably because it acts as increasing the learning rate vastly, which hinders the optimization. 
The best performance is achieved when this contrastive loss weight is a little lower than the segmentation losses $\mathcal{L}_{sup}$ and $\mathcal{L}_{pseudo}$ ($\lambda_{contr}=10^{-1}$).

The effect of the size (per-class) of our memory bank is studied in \Tab{tab:memory-size}. As expected, higher values lead to stronger performances, although from 256 up they tend to maintain similarly.
Because all the elements from the memory bank are used during the contrastive optimization \equ{eq:contrastive-loss} the larger the memory bank is, the more computation and memory it requires. Therefore, we select $256$ as our default value.

\Tab{tab:contrastive} studies the effect of the main components used in the proposed contrastive learning module. 
The base configuration of the module which includes our simplest implementation of the per-pixel contrastive learning using the memory bank,
still presents a performance gain compared to not using the contrastive learning module (57.4 mIoU from \ref{tab:ablation}).
Generating and selecting good quality prototypes is the most important factor. This is done both by the Feature Quality Filter (FQF), \ie checking that the feature leads to an accurate and confident prediction, and extracting them with the teacher network $f_{\xi}$. Then, using the class-specific attention $S_{c, \theta}$ to weight every sample (both from the memory bank and input sample) is also beneficial, acting as a learned sampling method.

\begin{table}[!tb]\label{tab:lr}
\begin{center}
\resizebox{0.49\textwidth}{!}{
\setlength{\tabcolsep}{5pt}
\begin{tabular}{l|cccccccc}
\hline 
$\lambda_{contr}$  & $10^4$ &  $10^2$ &  $10^1$ & $10^0$ &  $10^{-1}$ &$10^{-2}$ & $10^{-4}$  \\ 
 \hline
mIoU   & $50.3$ &  $51.4$   &  $54.8$   & $59.1$  & $59.4$ & $58.7$&  $57.6$  \\  

 \hline
\end{tabular}
 }
\end{center}

\caption{Effect of different values for the factor $\lambda_{contr}$ \equ{eq:approach} that weights the effect of the contrastive loss $\mathcal{L}_{\mathrm{contr}}$. 
Results on Cityscapes benchmark ($\frac{1}{30}$ available labels, Deeplabv2-ResNet101 COCO pre-trained).  
}
\end{table}

\begin{table}[!tb]
\label{tab:memory-size}
\begin{center}

\setlength{\tabcolsep}{5pt}
\begin{tabular}{l|cccccc}
\hline 
 $\psi$  &  $32$ & $64$ & $128$ &  $256$ &  $512$  \\ 
 \hline
 mIoU  &  $58.7$  & $58.9$ & $59.2$  &  $59.4$    &  $59.3$  \\
\hline
\end{tabular}

\end{center}

\caption{Effect of our memory bank size (features per-class), $\psi$. 
Results on Cityscapes benchmark ($\frac{1}{30}$ available labels, Deeplabv2-ResNet101 COCO pre-trained).}\vspace{-3mm} 
\end{table}

\begin{table}[!tb]\label{tab:contrastive}
\begin{center}
\setlength{\tabcolsep}{5pt}
\begin{tabular}{c|ccc|c}
\hline 
Base & $f_{\xi}$ & $S_{c, \theta}$ &  FQF &  mIoU \\
\hline
\checkmark & &     &   &  $58.3$  \\
\checkmark & \checkmark &    &   &  $58.7$ \\ 
\checkmark & &  \checkmark   &   &  $58.6$ \\
\checkmark & &  &   \checkmark    &  $59.0$ \\ 

\hline
\checkmark & \checkmark & \checkmark  & \checkmark  & $59.4$  \\
\hline

\end{tabular}

\small{\textit{$f_{\xi}$: Use teacher model $f_{\xi}$ to extract features instead of $f_{\theta}$} } \\  
\small{\textit{$S_{c, \theta}$: Use class-specific attention $S_{c, \theta}$  to weight every feature  } } \\  
\small{\textit{FQF: Feature Quality Filter for Memory Bank update}  } \\
\end{center}

\caption{Ablation study of our contrastive learning module main components. Results on Cityscapes benchmark ($\frac{1}{30}$ available labels, using Deeplabv2-ResNet101 COCO pre-trained).\vspace{-3mm} 
}
\end{table}

\vspace{-2mm} 
\paragraph{Future direction.} Our proposed approach could potentially be applied to other  semi-supervised tasks like object detection or instance segmentation. The straightforward way is to perform the proposed contrastive learning using the features from the semantic head of the detection or instance segmentation networks, i.e., the part of the network that outputs the semantic class of the object or instance.
The method is currently restricted by the number of classes and number of memory bank entries per class. A future step to solve this problem could be to cluster the feature vectors per class and save only cluster centers of the class features, similar to the very recent work from Zhang et. al  \cite{zhang2021prototypical} for domain adaptation based on prototypical learning.

\section{Conclusion}

This paper presents a novel approach for semi-supervised semantic segmentation. 
Our work shows the benefits of incorporating positive-only contrastive learning techniques to solve this semi-supervised task. The proposed contrastive learning module boosts the performance of semantic segmentation in these settings. 
Our new module contains a memory bank that is continuously updated with selected features from those produced by a teacher network from labeled data. These features are selected based on their quality and relevance for the contrastive learning.
Our student network is optimized for both labeled and unlabeled data to learn similar class-wise features to those in the memory bank. 
The use of contrastive learning at a pixel-level has been barely exploited and this work demonstrates the potential and benefits it brings to semi-supervised semantic segmentation and semi-supervised domain adaptation. Our results outperform state-of-the-art on several public benchmarks, with particularly significant improvements on the more challenging set-ups, \ie when the amount of available labeled data is low.

\section{Acknowledgments}
This work was partially funded by  FEDER/ Ministerio de Ciencia, Innovaci{\'o}n y Universidades/ Agencia Estatal de Investigaci{\'o}n/RTC-2017-6421-7, PGC2018-098817-A-I00 and PID2019-105390RB-I00, Arag{\'o}n regional government (DGA T45 17R/FSE) and the Office of Naval Research Global project ONRG-NICOP-N62909-19-1-2027.

{\small
\bibliographystyle{ieee_fullname}
\bibliography{egbib}
}

\end{document}